\pdfoutput=1

\documentclass[11pt]{article}

\usepackage{amsmath,amsfonts,bm}

\def\eqref#1{equation~\ref{#1}}

\def\1{\bm{1}}

\def\ra{{\textnormal{a}}}

\def\rr{{\textnormal{r}}}

\DeclareMathAlphabet{\mathsfit}{\encodingdefault}{\sfdefault}{m}{sl}
\SetMathAlphabet{\mathsfit}{bold}{\encodingdefault}{\sfdefault}{bx}{n}

\def\gD{{\mathcal{D}}}
\def\gE{{\mathcal{E}}}

\def\gR{{\mathcal{R}}}

\def\sS{{\mathbb{S}}}

\usepackage{EMNLP2023}

\usepackage{times}
\usepackage{latexsym}

\usepackage[T1]{fontenc}

\usepackage[utf8]{inputenc}

\usepackage{microtype}

\usepackage{inconsolata}

\usepackage{graphics}
\usepackage{graphicx}
\usepackage{amssymb}
\usepackage{amsmath}
\usepackage{mathrsfs}
\usepackage{amsfonts}
\usepackage{algorithmicx}
\usepackage[ruled, lined, linesnumbered, commentsnumbered, longend]{algorithm2e}

\usepackage{times}
\usepackage{latexsym}
\usepackage{multicol}
\usepackage{multirow}
\usepackage{array}
\usepackage{tabu}
\usepackage{booktabs}
\usepackage{enumitem}
\usepackage{makecell}
\usepackage{cleveref}

\newcommand\ee{$\textsc{EvEval}$\ }

\title{\ee: A Comprehensive Evaluation of Event Semantics for Large Language Models}
\author{Zhengwei Tao$^1$ ~Zhi Jin$^1$ ~Xiaoying Bai$^2$ ~ Haiyan Zhao$^1$ \\ ~{\bf Yanlin Feng$^1$} ~{\bf Jia Li$^1$} ~{\bf Wenpeng Hu$^1$} \\
        $^1$Peking University, $^2$Advanced Institute of Big Data \\ \texttt{tttzw@pku.stu.edu.cn},  ~\texttt{\{zhijin,zhhy.sei,fengyanlin,lijiaa,wenpeng.hu\}@pku.edu.cn} \\ \texttt{baixy@aibd.ac.cn}}

\begin{document}
\maketitle
\begin{abstract}
Events serve as fundamental units of occurrence within various contexts. The processing of event semantics in textual information forms the basis of numerous natural language processing (NLP) applications. Recent studies have begun leveraging large language models (LLMs) to address event semantic processing. However, the extent that LLMs can effectively tackle these challenges remains uncertain. Furthermore, the lack of a comprehensive evaluation framework for event semantic processing poses a significant challenge in evaluating these capabilities. In this paper, we propose an overarching framework for event semantic processing, encompassing understanding, reasoning, and prediction, along with their fine-grained aspects. To comprehensively evaluate the event semantic processing abilities of models, we introduce a novel benchmark called \ee. We collect 8 datasets cover all aspects of event semantic processing. Extensive experiments are conducted on \ee, leading to several noteworthy findings based on the obtained results.

\end{abstract}


\section{Introduction}
Events are situations that happen or occur constituting the fundamental semantic foundation encompassing the meanings of \textit{Activities}, \textit{Accomplishments}, \textit{Achievements}, and \textit{State}~\cite{vendler1957verbs}. Event semantics are meanings of events to textual data. Processing event semantics underlies NLP applications such as document retrieval~\cite{rudnik2019searching}, recommendation system~\cite{yang2020temporal}, and question answering~\cite{souza2020event}. As a crucial and wildly-used technical library, it mainly requires machines to be capable of understanding~\cite{pedinotti2021did}, reasoning~\cite{han2021ester}, and predicting~\cite{zhao2021event} of the event semantics. Recently, there has been an increasing focus on tackling the issues related to event semantics in neural models and enhancing their abilities.

\begin{figure}[!tb]
    \centering
    \includegraphics[width=1\columnwidth]{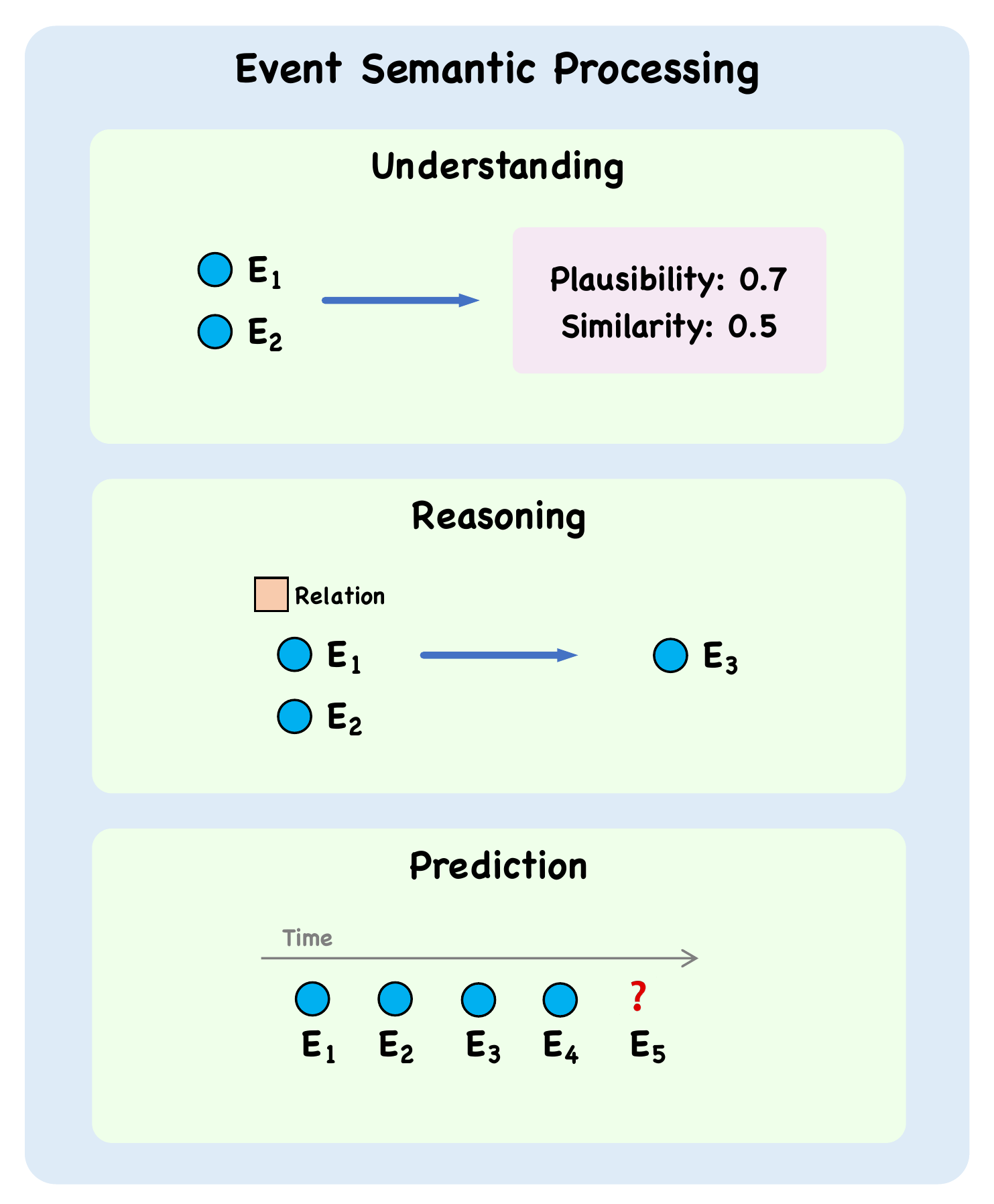}
    \caption{Event Semantic Processing consists of understanding, reasoning, and prediction. }
    \label{fig:ESP}
\end{figure}

In the recent past, various large language models (LLMs) like LLAMA~\cite{touvron2023llama}, BLOOM series~\cite{scao2022bloom, muennighoff2022crosslingual}, and GPT series~\cite{brown2020language} have demonstrated exceptional accomplishments in various natural language processing (NLP) tasks, such as inference, contextual comprehension, chain-of-thought reasoning~\cite{wei2022chain}, and demonstrating plausible explanations~\cite{lampinen2022can}. Nonetheless, the potential and capability of large language models for event semantic processing still remains unclear. Exploring and understanding the merits of how an LLM processes the event semantics is key to adapting the great pre-training ability to event semantics processing. Furthermore, the absence of a comprehensive framework for event semantic processing makes the assessment of its abilities a challenging task.

In this paper, we initially establish a comprehensive hierarchy of abilities for event semantic processing as shown in Figure~\ref{fig:ESP}. The primary expectation of humans is for machines to comprehend intra- and inter-events accurately and consistently, similar to how we as humans do~\cite{pedinotti2021did}. With a more nuanced comprehension, we delve deeper into more inquiries such as causes, outcomes, and temporal aspects of events~\cite{han2021ester}. Expanding on existing relational reasoning, our ultimate objective revolves around predicting future events based on different representations~\cite{zhao2021event}.
Motivated by this, we introduce \ee, a new benchmark designed to evaluate the capacities of processing event semantics for the LLM. In \ee, we include 8 datasets covering \textbf{understanding}, \textbf{reasoning}, and \textbf{prediction} about the event semantics. In the part of event semantic understanding, \ee aims to verify the ability to understand intra- and inter-events. Understanding an atomic event and its relation to others is the basis of event semantic processing. Consequently, \ee provides evidence of the reasoning capabilities of LLMs within event-centric dimensions which encompass causality~\cite{du2022care}, temporality~\cite{zhou2020temporal}, counterfactual~\cite{qin2019counterfactual}, and intent~\cite{sap2019socialiqa}. These dimensions reach the main aspects of the event semantic reasoning and are applicable to wild sorts of NLP tasks. Building upon the aforementioned abilities, event semantic prediction requires the anticipation of future events by leveraging present occurrences. It involves two types: script-based and story-based. The script-based one focuses on prototypical events, which requires prior event semantical knowledge. Conversely, the story-based counterpart necessitates the comprehension of contextual event information. 

Furthermore, \ee aspires to identify additional features during the event semantic processing by LLMs. We raise several key points: 1) How does chain-of-thought~(CoT)~\cite{wei2022chain} improve event semantic processing? 2) Can LLM only process event semantics in natural language representation?  To probe such points, we include additional two tasks of CoT and event representations. These tasks are designed for additional investigations to gain a deeper understanding of event semantic processing.

We then conduct a series of experiments on \ee with several LLMs. In addition to the primary findings, we also investigate several attributes related to event semantic processing. Our findings suggest that 1) While LLMs exhibit comprehension of individual events, their ability to grasp the semantic similarity between events is limited.
2) LLMs showcase strong reasoning capabilities in causal and intentional relations, but their performance in other relation types is relatively inferior.
3) LLMs demonstrate improved forecasting of future events with greater contextual information.
4) The extent of utilization of event knowledge, context, and patterns may be essential for CoT in event semantic processing.
5) Structural event representation shows comparable performance to natural language in the context of event processing. We summarize our contributions:
\begin{itemize}[topsep=0pt]
\setlength{\itemsep}{1pt}
\setlength{\parskip}{0pt}
\setlength{\parsep}{-1pt}
\setlength{\leftmargin}{-1pt}
\item[$\bullet$] We propose a comprehensive ability hierarchy of event semantic processing. Based on that, we introduce \ee, a comprehensive benchmark for evaluating the capacities of LLM on event semantic processing. \ee
\ consists of our induced aspects of ability covering event semantic understanding, reasoning, and prediction.
    
\item[$\bullet$] We design further experiments to probe the characteristics of event semantic processing.
    
\item[$\bullet$] Through the identification of the strengths and weaknesses of LLMs in event semantic processing, we offer valuable insights to guide the advancement of more sophisticated language models capable of effectively addressing complex event semantic tasks.

\end{itemize}
\section{Preliminaries}
\subsection{Event Definition}
An Event is something that happens involving participants~\cite{doddington2004automatic}. An event may have correlations with others. Formally, 

$\gE$ is an event consists of several participants or arguments. $\ra_i$ is the $i^{th}$ argument, $\rr_i\in \sS^{\rr}$ is the role of $\ra_i$. The event $\gE$ can have a relation $\gR\in \sS^{\gR}$ with other events. $\sS^{\rr}$ is the set of all argument roles and $\sS^{\gR}$ is the set of event relations.

Conventionally, $\sS^{\rr}$ includes \texttt{predicate}~(also called \texttt{trigger}), \texttt{subject}, \texttt{object}, \texttt{time}, \texttt{location}, etc.. $\sS^{\gR}$ consists of \texttt{causal}, \texttt{temporal}, \texttt{counterfactual}, \texttt{intentional}, etc.. 

\subsection{Event Semantic Processing}
Presently, there is an increasing abundance of textual information sources containing extensive descriptions of events. These events are represented in a natural language known as event mentions which can be words, phrases, and sentences~\footnote{Without loss of generosity, we also use $\gE$ to denote the event mentions in natural language.}. Event semantic processing aims to comprehend the meanings of events with their participants and inter-relations embedded in the textual data.

From the macro view of event semantic processing, human mainly demands machines to understand events correctly and consistently the same as our human beings~\cite{pedinotti2021did}. Based on the sophisticated understanding, we further question for causes, results, or temporality of events~\cite{han2021ester}. Building upon prior relational reasoning, our ultimate inquiry revolves around predicting future occurrences~\cite{zhao2021event}. 

Based on existing research of interests and the real application, we are motivated to induce the abilities demanded by LLMs to process event semantics as understanding, reasoning, and prediction. We introduce each aspect as follows:
\begin{itemize}[topsep=0pt]
\setlength{\itemsep}{1pt}
\setlength{\parskip}{0pt}
\setlength{\parsep}{-1pt}
\setlength{\leftmargin}{-1pt}
\item[] \textit{Understanding.} A large language model is required to accurately comprehend the meaning of an event and ascertain its plausibility and likelihood of occurrence.
    
\item[] \textit{Reasoning.} A large language model should be capable of reason about some factors of an event. We induce the factors as causality, temporality, counterfactual, and intent.
    
\item[] \textit{Prediction.} A large language model should predict future events according to its analysis of current situations. A model should have the prototypical logic between events. 

\end{itemize}

\section{EvEval Benmark}
\label{ee}
In this section, we provide detailed explanations and elaborations on the \ee benchmark. The overview of the \ee is in Table~\ref{overview}

\begin{table*}[!t]
\centering
\small
\setlength{\tabcolsep}{1mm}{\begin{tabular}{llp{12.5cm}}
\toprule
\multicolumn{1}{l}{\textbf{Dataset}} & \multicolumn{1}{l}{\textbf{Task}} & \multicolumn{1}{l}{\textbf{Example}}\\\midrule\midrule
\multicolumn{3}{c}{\textsc{Understanding}} \\ \midrule
 \multirow{2}{*}{\texttt{DTFit}} &\multirow{2}{*}{\texttt{Intra}}& \textbf{Event1:} The swimmer wore the goggles during the dinner.\textbf{Event2:} The swimmer wore the goggles during the competition. \textbf{Question:} Which one is a more plausible event?~\textbf{Answer:} Event1. \\ \midrule
 
\multirow{2}{*}{\texttt{HardSim}} & \multirow{2}{*}{\texttt{Inter}} & \textbf{Event:} They begin negotiations.\textbf{Question:} Which one of the choices is more semantically similar to the event above? \textbf{A.} People start talks.~\textbf{B.} They begin career.~\textbf{Answer:} A \\ \midrule
 
 \multicolumn{3}{c}{\textsc{Reasoning}} \\ \midrule
  \multirow{3}{*}{\texttt{ECARE}} & \multirow{3}{*}{\texttt{Causal}} & \textbf{Event:} The man devoted himself to journalism.\textbf{Question:} Which one of the choices is result of the event above? \textbf{A.} He mastered the popular language used in pursuit of truth.~\textbf{B.} He became fatter.~\textbf{Answer:} A \\ \midrule

 \multirow{4}{*}{\texttt{TRACIE}} & \multirow{4}{*}{\texttt{Temporal}} & \textbf{Context:} Margaret was walking through town. She noticed a store window with an ad for a family shelter. The pictures of kids really struck a chord in her heart. She decided to donate. She walked inside and began the process. \textbf{Question:} Is it true that she felt guilty starts after she saw an ad for a family shelter?~\textbf{Answer:} True \\ \midrule

 \multirow{8}{*}{\texttt{TIMETRAVEL}} & \multirow{8}{*}{\texttt{Count.}} & \textbf{Premise:} The soccer game was tied 3 to 3 and there was a minute left to play. \\ 
 && \textbf{beginning:} Julie had never scored a goal yet, but knew today would be her day. \\
  &&\textbf{Ending:} Ashley passed her the ball and this was chance. She kicked as hard as she could, and the ball soared into the net. Julie's first goal won the game. \\
 &&\textbf{Edited beginning:} Julie was eagerly watching the game in the stands. \\
 &&\textbf{Question:} What will be the edited ending? \\
 &&\textbf{Edited ending:} Ashley passed the ball and this their was chance. She kicked as hard as she could, and the ball soared into the net. Julie's team won the game.\\\midrule

 \multirow{3}{*}{\texttt{SocialIQA}} & \multirow{3}{*}{\texttt{Int.}} & \textbf{Event:} Kendall watched baseball with his friends after working hard all week at his job. \textbf{Question:} What will Kendall want to do next? \textbf{A.} Avoid talking to his friends.~\textbf{B.} Cheer his team with his friends.~\textbf{C.} Needed to please her boss.~\textbf{Answer:} B\\ \midrule

 \multicolumn{3}{c}{\textsc{Prediction}} \\ \midrule
 \multirow{4}{*}{\texttt{MCNC}} & \multirow{4}{*}{\texttt{Script}} & \textbf{Events:} Canada agrees. Canada pays. Canada represents a commitment. Canada fish out. Canada slapped the fee. Canada sent boat strait. Canada responded. Canada lifting fee. \textbf{Question:} What will the next event?\textbf{A.} Canada votes player.~\textbf{B.} Canada seems.~\textbf{C.} Canada advises.~\textbf{D.} Canada be ok~\textbf{E.} Canada conducts the review.~\textbf{Answer:} E \\\midrule

 \multirow{4}{*}{\texttt{SCT}} & \multirow{4}{*}{\texttt{Story}} & \textbf{Events:} Oliver was nervous about his wedding. He was worried that he would stutter during the vows. When the time came, he took a deep breath and began to speak. He stuttered, but his wife smiled and hugged him and he was okay. \textbf{Question:} What will the next event? \textbf{A.} Oliver decided to not get married.~\textbf{B.} Oliver was so grateful for his wife's love.~\textbf{Answer:} B \\\midrule
 
\end{tabular}}
\caption{Overview of \ee. We showcase one datum of each dataset. \texttt{Count.} and \texttt{Int.} stand for counterfactual and intentional reasoning. }
\label{overview}
\end{table*}

In order to comprehensively evaluate a model's abilities to process event semantics, as we have induced, we collect 8 datasets covering event semantic understanding, reasoning, and prediction.

\subsection{Understanding}
The understanding part of \ee aims to testify whether a model can comprehend the meaning of the event correctly as a human does and can judge the possibility of an event happening. We design two tasks for this purpose: intra-event and inter-event understanding.

\begin{itemize}[leftmargin=*]
\setlength{\itemsep}{1pt}
\setlength{\parskip}{0pt}
\setlength{\parsep}{-1pt}
\item[] \textbf{Intra}~[\texttt{DTFit}]. Given two events $\gE_1=\{(\rr_i, \ra_i)| i\in{[1, k_1]}\}$ and $\gE_2=\{(\rr_i, \ra_i)| i\in{[1, k_2]}\}$, the model should pick out the more plausible event. This task evaluates the model's ability to comprehend the roles and participants involved in events, as well as its capacity to generate plausible event representations. We use accuracy for the metric of this task.

For the Intra-Event task, we use the \texttt{DTFit} dataset~\cite{pedinotti2021did}. The dataset consists of pairs of events that have a single differing argument, representing either a typical or atypical filler for a specific role in the described event. To gather typical judgments, the authors obtained assessments from native English speakers regarding the frequency of occurrence for each described event. 

\item[] \textbf{Inter}~[\texttt{HardSim}]. Given an event $\gE$ and two candidate events $\gE_1$ and $\gE_2$, the model should find which one of the candidate events is more semantically similar to $\gE$. This task evaluates the ability to distinguish the event semantics. We use accuracy for the metric of this task.

We adopt \texttt{HardSim}~\cite{ding2019event} for the Inter-Event task. Each data has an event and two choices. One with similar semantics but has very little lexical overlap while another one with dissimilar semantics but has high lexical overlap. 
 
\end{itemize}

\subsection{Reasoning}
In the reasoning part, \ee evaluates a model's reasoning ability on event semantics. We meticulously categorize this evaluation into multiple tasks to ensure a comprehensive assessment.

\begin{itemize}[leftmargin=*]
\setlength{\itemsep}{1pt}
\setlength{\parskip}{0pt}
\setlength{\parsep}{-1pt}
\item[] \textbf{Causal}~[\texttt{ECARE}]. Given an event $\gE$ and a target relation $\gR$ which can be either cause or effect, the model should pick out the correct event from the candidate events according to $\gR$. We use accuracy for the metric of this task.

We choose \texttt{ECARE}~\cite{du2022care} dataset. This dataset is designed for explainable causal reasoning and comprises causal questions, accompanied by a vast collection of unique conceptual explanations that facilitate a profound comprehension of causal facts.

\item[] \textbf{Temporal}~[\texttt{TRACIE}]. Given an event $\gE$ and a target relation $\gR$ which can be either before or after, the model should pick out the correct event from the candidate events according to $\gR$. We use accuracy for the metric of this task.

We use \texttt{TRACIE}~\cite{zhou2020temporal}. The primary emphasis of this dataset revolves around temporal relationships concerning implicit events present in short stories. \texttt{TRACIE} dataset incorporates meticulously annotated start and end time queries, which serve as a comprehensive evaluation of a system's grasp of complete temporal closure (i.e., both start and end times) pertaining to events.

\item[] \textbf{Counterfactual}~[\texttt{TIMETRAVEL}]. Given a premise event $\gE^{p}$, a beginning event $\gE^{b}$, and an origin story endding $\gD^{o}$. The model should be able to edit the $\gD^{o}$ to $\gD^{c}$ according to a counterfactual beginning event $\gE^{c}$. We use the bleu and rouge scores~\cite{papineni2002bleu,lin2004rouge} and for the metric of this task.

We adopt \texttt{TIMETRAVEL}~\cite{qin2019counterfactual} dataset. It introduces the concept of Counterfactual Story Rewriting, which involves the task of making minimal revisions to an original story in order to align it with a provided counterfactual event.

\item[] \textbf{Intent}~[\texttt{SocialIQA}]. Given an event $\gE$, the model should pick out the correct event from the candidate events most reflect the agent's intent. We use accuracy for the metric of this task.

We include \texttt{SocialIQA}~\cite{sap2019socialiqa} dataset. It comprises a collection of multiple-choice questions designed to assess emotional and social intelligence across diverse everyday scenarios. We here only use the intentional part of \texttt{SocialIQA}. Specifically, we filter the dataset by only using questions like "What does he want to do?".

\end{itemize}

\subsection{Prediction}
In the prediction part, \ee verifies a model for predicting future events based on happened events. We include two tasks:

\begin{itemize}[leftmargin=*]
\setlength{\itemsep}{1pt}
\setlength{\parskip}{0pt}
\setlength{\parsep}{-1pt}

\item[] \textbf{Script}~[\texttt{MCNC}]. Given a chronological sequence of structured script events $\gE_n=\{(\rr_i, \ra_i)\}, n\in{[1, N]}$ all with subject, trigger, and some object. The task is to select the correct future event from the given candidate events. We use accuracy for the metric of this task.

We utilize \texttt{MCNC}~\cite{granroth2016happens} dataset. It is closely related to the narrative cloze task but better suited to comparing systems’ usefulness as a component in a narrative prediction.

\item[] \textbf{Story}~[\texttt{SCT}]. Given a chronological sequence of story events $\gE_n, n\in{[1, N]}$ in natural language. The task is to select the correct future event from the given candidate events. We use accuracy for the metric of this task.

We use \texttt{SCT}~\cite{mostafazadeh2016corpus} dataset. It encompasses a diverse range of causal and temporal commonsense relations between everyday events, and it serves as a high-quality compilation of stories from everyday life.

\end{itemize}

\subsection{Chain-of-Thought}
Chain-of-Thought~(CoT)~\cite{wei2022chain} is an enhanced prompting approach designed to enhance the proficiency of LLMs in handling intricate reasoning tasks, including arithmetic reasoning~\cite{cobbe2021training}, commonsense reasoning~\cite{wei2022chain}, and symbolic reasoning~\cite{wei2022chain}, thereby improving their overall performance. \ee consists of evaluating CoT prompting since event semantic processing is a group of cognitive skills. In this part, \ee tests various CoT methods on all tasks.

\subsection{Event Representation}
Event representation learning has garnered significant attention in recent times, focusing on discovering formats that can more effectively embed the semantics of events~\cite{chambers2008unsupervised,chambers2009unsupervised,gao2022improving}. As the demand for LLMs continues to rise, there is a growing interest in event representations specifically tailored for LLMs. On the one hand, within the realm of embodied AI, researchers are exploring ways to leverage LLMs to assist in controlling robots to accomplish various events~\cite{yuan2023plan4mc,liang2022code,driess2023palm}. On the other hand, it can conduct event simulations~\cite{park2023generative}. It naturally posts a question: \emph{what are better representations of events when interacting with LLM?}

Existing ways include natural language~\cite{yuan2023plan4mc, driess2023palm}, structural forms~\cite{wang2022code4struct,liang2022code}. Natural language representations align more consistently with LLMs, whereas structural forms offer increased parseability and determinism.

Motivated by this, we include a representation evaluation in \ee. This task testifies what representations are better for event semantic processing. We use three datasets: \texttt{DTFit}, \texttt{HardExt}, and \texttt{MCNC} in this part. We tag argument roles for each data of these datasets. The task requires methods to compose better event representations to achieve higher scores.


\section{Evaluation on \ee}
In this section, we comprehensively evaluate several LLMs on \ee according to aspects mentioned in Section~\ref{ee} and analyze the results. We first briefly introduce the evaluated LLMs in~\ref{llm}. Then we report results of understanding, reasoning, and prediction respectively in~\ref{ru}, \ref{rrea}, and \ref{rp}. After that, we evaluate the influence of CoT and representation on event semantic processing in~\Cref{rc,rrep}. 

\subsection{Evaluated LLMs}
\label{llm}
We introduce the LLMs as follows.

\noindent\texttt{ChatGPT}. A large language model trained by OpenAI\footnote{https://chat.openai.com/}. It utilizes a transformer architecture and first trains on unsupervised learning. The model is then fine-tuned using a combination of supervised and reinforcement learning methods to improve its performance. 

\noindent\texttt{BLOOM}~\cite{scao2022bloom}. BLOOM is an LLM that has been trained extensively on massive text datasets using substantial computational resources. It excels at generating coherent text in 46 languages and 13 programming languages, often indistinguishable from human-written text. 

\noindent\texttt{BLOOMZ}~\cite{muennighoff2022crosslingual}. This is a model capable of following human instructions in dozens of languages zero-shot. It is finetuned based on BLOOM by using cross-lingual task mixture datasets~(xP3).

\noindent\texttt{Flan-T5}~\cite{chung2022scaling}. This is a model finetuned on T5 with a scaling number of datasets which achieves strong few-shot performance even compared to much larger models. We use the xxl version which has 11B parameters.

\subsection{Results on Understanding}
\label{ru}
We test models in the understanding part with the same prompt. The results are shown in Table~\ref{eval}. 

\noindent\emph{Findings 1. LLMs understand individual events, but not the semantic similarity between events.} 

We find models accomplish event semantic understanding conditionally. Most LLMs exhibit an ability to grasp intra-event semantics but struggle with comprehending inter-event semantics. Taking ChatGPT as an example, ChatGPT achieves 91.43 accuracies on \texttt{DTFit} but only 65.44 on \texttt{HardExt}. This demonstrates to most LLMs what a plausible event is like and understands event arguments and their roles. It indicates that language modeling training is sufficient to learn individual event semantics. However, inter-event semantics is hard to learn which demonstrates that there may be a gap between understanding individual events and inter-event semantics. There may exist more steps such as semantics alignment. Neither unsupervised training nor instruction-tuning can provide enough such alignments.

\begin{table*}[!t]
\centering
\small
\setlength{\tabcolsep}{2.8mm}{\begin{tabular}{lcccccccc}

\toprule
& \multicolumn{2}{c}{\textsc{Understanding}} & \multicolumn{4}{c}{\textsc{Reasoning}}& \multicolumn{2}{c}{\textsc{Prediction}} \\
\cmidrule(lr){2-3}\cmidrule(lr){4-7}\cmidrule(lr){8-9}
 & \multicolumn{4}{c}{ACC} & BLEU4/Rouge-L & \multicolumn{3}{c}{ACC} \\
\cmidrule(lr){2-5}\cmidrule(lr){6-6}\cmidrule(lr){7-9} 
 & \textbf{Intra}& \textbf{Inter}& \textbf{Causal}& \textbf{Temporal}& \textbf{Counterfactual}& \textbf{Intentional}& \textbf{Script}& \textbf{Story} \\
 & \texttt{DTFit}& \texttt{HardExt}& \texttt{ECARE}& \texttt{TRACIE}& \texttt{TIMETRAVEL}& \texttt{SocialIQA}& \texttt{MCNC}& \texttt{SCT} \\
 \midrule
 \multicolumn{9}{c}{\textsc{Finetune}} \\
 \midrule
 SOTA &-&-&78.00&80.60&87.18~/~80.74& - & 64.61&91.33\\
 LM &-&-&70.73&78.40&82.91~/~76.44& 72.56 & 61.53&87.10\\
 \midrule
 \multicolumn{9}{c}{\textsc{Zero-Shot}} \\
 \midrule
 ChatGPT & 91.43&65.44&78.28&54.60&1.72~/~24.97&64.80&32.70&89.85 \\
 BLOOM & 47.89&43.33&56.14&49.87&17.99~/~46.43&37.73&21.60&63.23 \\
 BLOOMZ & 88.34&81.00&79.83&54.47&-&76.71&38.50&96.95 \\
 Flan-T5 & 50.28&49.22&63.27&50.92&0.18~/~8.83&40.07&21.60&80.71 \\
 \bottomrule
\end{tabular}}
\caption{Evaluation results on \ee of understanding~(\ref{ru}), reasoning~(\ref{rrea}) and prediction~(\ref{rp}). LM stands for language models where it's BART~\cite{lewis2019bart} for TIMETRAVEL and RoBerta~\cite{liu2019roberta} for others.}
\label{eval}
\end{table*}

\subsection{Results on Reasoning}
\label{rrea}

We use the same prompt for all models in the reasoning part. We list results in Table~\ref{eval}.

\noindent\emph{Findings 2. LLMs demonstrate proficient reasoning abilities in causal and intentional relations, but their performance in other types of relations is comparatively weaker.} 

ChatGPT achieves 78.28 accuracy which is even higher than SOTA performance in causal reasoning. BLOOMZ gains a 76.71 accuracy score which is better than finetuned Roberta. However, it falls far behind SOTA or finetuning performance in Temporal, Counterfactual reasonings. This may probably suggest that the event causality and intent patterns are more available in the training data while others are rarer. Another explanation for this phenomenon is that event causality semantics may be easier to learn and more determined. Learning other relations such as event temporality needs more context information and knowledge since it is more subjected to specific situations~\cite{ghosh2022pasta}.

\subsection{Results on Prediction}
\label{rp}

In this part, we test LLMs on prediction tasks. The results are shown in Table~\ref{eval}.

\noindent\emph{Findings 3. The more context, the better LLMs can forecast future events.}

As the results show, we find LLMs can complete story-based event prediction well, but almost fail in the script-based one. We analyze this could be due to the story-based prediction having more context information in its wording. Nevertheless, the prediction based on scripts only includes primary arguments, resulting in greater variability. This is consistent with human behavior of predicting future events. The progression of events is influenced not only by logical knowledge of the events but also by situational factors~\cite{li2021future}.

\begin{table}[!t]
\centering
\small
\setlength{\tabcolsep}{3.15mm}{\begin{tabular}{llcc}
\toprule

&& w.o. CoT & w.t. CoT \\
\midrule
\multicolumn{4}{c}{\textsc{Understanding}} \\
\midrule
\texttt{DTFit}& \textbf{Intra}&91.43&81.74\\
\texttt{HardExt}& \textbf{Inter}&65.44&67.33\\
\midrule
\multicolumn{4}{c}{\textsc{Reasoning}} \\
\midrule
\texttt{ECARE}& \textbf{Causal}&82.36&72.75\\
\texttt{TRACIE}& \textbf{Temporal}&54.60&52.11 \\
\texttt{SocialIQA}& \textbf{Intentional}&64.80&65.70 \\
\midrule
\multicolumn{4}{c}{\textsc{Prediction}} \\
\midrule
\texttt{MCNC}& \textbf{Script}&32.70&24.40\\
\texttt{DTFit}& \textbf{Story}&89.85&87.55\\

\bottomrule

\end{tabular}}
\caption{ChatGPT Chain-of-Thought performances on \ee.}
\label{cot}
\end{table}

\subsection{Results on CoT}
\label{rc}

\begin{table*}[!t]
    \centering
    \small
    \begin{tabular}{p{7cm}p{8.3cm}c}
         \toprule
         Question & CoT Output\\
         \hline
         Event: Now it's time to remove the drugs from the patient's body. \newline
         Question: What is the effect of the event? \newline
         Choices: A. The doctor uses the method of plasmapheresis to treat him. B. Various elimination processes were adopted to achieve that. \newline
         Answer: Let's think step by step. \newline\newline
         Event: The man devoted himself to journalism. \newline
         Question: What is the effect of the event? \newline
         Choices: A. He mastered the popular language used in pursuit of truth.  B. He became fatter. Answer: Let's think step by step.
         & The event mentioned is "removing the drugs from the patient's body". The question is asking about the effect of this event. Option A suggests that the effect is the doctor using the method of plasmapheresis to treat the patient, but this is not necessarily the case. Option B, which suggests that various elimination processes were adopted to achieve this, is a more general and accurate answer. Therefore, the answer is B. \newline \newline
         There is no indication in the event that the man learned a new language or became more skilled in a specific language, so option A is incorrect. Similarly, there is no direct association between journalism and weight gain, so option B is also incorrect. Therefore, the answer is neither A nor B.\\
         \bottomrule
    \end{tabular}
    \caption{Two cases of CoT of \texttt{ECARE} dataset. The correct labels for both cases are B and A.}
    \label{cot case}
\end{table*}

We testify to the performances of ChatGPT on Chain-of-Thought evaluations. For all tasks, we implement CoT by adding the prompt "Let's think step by step."~\cite{wei2022chain}. However, the CoT outputs may be distinctly formed making them hard to parse the answer. Therefore, we call ChatGPT to parse each answer by adding the following prompt to the original question and CoT output: "\textit{[Question] [CoT] Therefore, the answer is}". 

\noindent\emph{Findings 4. CoT for event semantic processing may need better utilization of event knowledge, context, and patterns.}

The CoT results are shown in Table~\ref{cot}. In general, adding CoT obtains little gains for event semantic reasoning. Most of the tasks would drop if adding CoT prompting, except for \texttt{HardSim} and \texttt{SocialIQA}. We analyze this phenomenon and consider the multiple possible reasons. We show some cases in Table~\ref{cot case}.

Firstly, event semantic processing may not be tasks with much more reasoning steps. CoT excels in multi-step reasoning tasks such as arithmetic reasoning, and symbolic reasoning~\cite{wei2022chain}. However, event semantic processing requires more event knowledge. As the first example in Table~\ref{cot case}, the CoT output doesn't have a reasoning chain but one-hop explanations of each choice. Since the model can explain both choices well, it returns the correct answer. Secondly, event semantic processing demands a model to generate more useful patterns of event logic. Recent studies shows that the text assists LLMs in generating valuable patterns, while these patterns aid LLMs in comprehending tasks and generating texts that contribute to their solution~\cite{madaan2022text}. It accounts if the model can generate such patterns of events. As the second example in Tabel~\ref{cot case}, the model can't such patterns or context for choices, therefore, it fails to answer the question. Combining these two analyses, an effective CoT prompt for event semantic processing should enable generating of context, knowledge, and patterns of events.

\subsection{Results of Representations}
\label{rrep}
\begin{table*}[!t]
    \centering
    \small
    \setlength{\tabcolsep}{0.1mm}{\begin{tabular}{p{6cm}p{10cm}}
         \toprule
         NL & JSON \\
         \midrule
         The cheerleader shook the flag in the garage. & \{SUBJECT: cheerleader, VERB: shake, OBJECT: flag, LOCATION: garage\} \\
         \\
         The carver built the sculpture with the chisel. & \{SUBJECT: carver, VERB: build, OBJECT: sculpture, INSTRUMENT: chisel\}\\
         \bottomrule
    \end{tabular}}
    \caption{Representaions of natural language~(NL) and JSON. Data are sampled from \texttt{DTFit}.}
    \label{repr case}
\end{table*}

In this section, we evaluate different event representations for ChatGPT. We compare two kinds of common representations. One is natural language, the second is JSON format. We show two formats in Table~\ref{repr case}. The structural JSON format is more understandable for robot systems while losing language coherence. The balance between structural and language coherence is a trade-off in event representations.

\noindent\emph{Findings 5. Structural event representation performs comparatively to natural language.}

We show results in Table~\ref{rrep}. Two representations perform comparatively by ChatGPT on three datasets. This result indicates again that the model understands arguments in natural language form which is consistent with the findings in event semantic understanding. Besides, it shows that LLM can directly interact with structural representations. This observation can provide valuable insights for research fields such as embodied AI, indicating that the design of robot-readable prompts can also be applicable to large language models. Lastly, there is a significant opportunity for exploring enhanced event representations for LLM, considering that even a basic JSON format yields satisfactory results. For instance, representations such as code may yield even greater benefits.~\cite{wang2022code4struct}.

\begin{table}[!t]
\centering
\small
\setlength{\tabcolsep}{8.5mm}{\begin{tabular}{lcc}
\toprule
Dataset & NL & JSON \\
\midrule
\texttt{DTFit} & 91.43& 92.42\\
\texttt{HardExt} & 65.44& 66.00\\
\texttt{MCNC} & 32.70& 32.26\\

\bottomrule

\end{tabular}}
\caption{ChatGPT performances of different representations.}
\label{repr}
\end{table}

\section{Characteristics Exploration}
In this section, we explore several characteristics of event semantic processing.

\subsection{Event Context}
We explore how the context will influence the performance.

\noindent\textbf{In-Context Learning.} In-context learning~(ICL) employs a structured natural language prompt, which includes a few task examples as demonstrations~\cite{brown2020language}. LLMs have the ability to recognize and execute new tasks without the need for explicit gradient updates, leveraging task demonstrations as a basis. In this part, we inspect how the demonstration context influences the performances. We conduct in-context learning on \ee based on ChatGPT. We use randomly sample 8 demonstrations for all datasets. The results are shown in Table~\ref{icl}.

We find that the ICL effect significantly enhances performance across all tasks. This indicates that the model is able to learn event knowledge and logic from demonstrations. Nevertheless, there is still room for enhancing event-oriented ICL algorithms.

\begin{table}[!t]
\centering
\small
\setlength{\tabcolsep}{2.3mm}{\begin{tabular}{llcc}
\toprule

$K$&& 0 & 8 \\
\midrule
\multicolumn{4}{c}{\textsc{Understanding}} \\
\midrule
\texttt{DTFit}& \textbf{Intra}&91.43&95.08\\
\texttt{HardExt}& \textbf{Inter}&65.44&75.33\\
\midrule
\multicolumn{4}{c}{\textsc{Reasoning}} \\
\midrule
\texttt{ECARE}& \textbf{Causal}&78.28&82.36\\
\texttt{TRACIE}& \textbf{Temporal}&54.60&65.00 \\
\texttt{TIMETRAVEL} & \textbf{Count.} &1.92/24.97& 25.17/45.3\\

\texttt{SocialIQA}& \textbf{Intentional}&64.80&69.68 \\
\midrule
\multicolumn{4}{c}{\textsc{Prediction}} \\
\midrule
\texttt{MCNC}& \textbf{Script}&32.70&40.50\\
\texttt{DTFit}& \textbf{Story}&89.85&95.88\\

\bottomrule

\end{tabular}}
\caption{ChatGPT in-context learning performances on \ee. $K$ stands for the number of demonstrations. \textbf{Count.} stands for counterfactual relation.}
\label{icl}
\end{table}

\section{Related Works}
\subsection{Benchmarks for LLM}
In order to assess the efficacy and superiority of LLMs, a multitude of tasks and benchmarks have been employed to conduct empirical evaluations and analyses. \citet{wang2018glue} presented GLUE, a comprehensive platform and resource collection designed for the evaluation and analysis of natural language understanding systems. \citet{wang2019superglue} introduced SuperGLUE, a novel benchmark designed to evaluate the capabilities of general-purpose language understanding systems. SuperGLUE enhances the existing GLUE benchmark by incorporating a fresh set of challenging natural language understanding tasks. \citet{lu2021codexglue} proposed CodeXGLUE, a benchmark with sorts of tasks about code processing, to facilitate the advancement of models capable of addressing diverse program understanding and generation challenges. \citet{li2023api} presented API-Bank, a benchmark explicitly crafted for Tool-Augmented LLMs. It encompasses a comprehensive workflow for Tool-Augmented LLMs, 53 widely utilized API tools, and 264 annotated dialogues, encompassing a total of 568 API calls.

\subsection{Evaluations for LLM}
Since its inception, LLM has consistently exhibited impressive capabilities. Evaluating the capacities of LLM is the foundation of using and improving them. \citet{bang2023multitask} performed a comprehensive technical assessment of ChatGPT by employing 23 datasets that span across 8 distinct NLP application tasks. \citet{bian2023chatgpt} conducted an investigation into the commonsense abilities of large language models and discovered that ChatGPT exhibits substantial knowledge but lacks experience as a problem solver. \citet{gao2023exploring,wei2023zero,li2023evaluating} explored the information extraction abilities of ChatGPT. They found that ChatGPT exhibits the problem of excessive confidence in its predictions. Additionally, in the majority of instances, ChatGPT demonstrates a notable level of fidelity to the original text. \citet{yuan2023zero} evaluated the ability to solve event relation extraction.

\subsection{Event Semantic Processing}
\noindent\textbf{Understanding} Event semantic understanding aims to understand the meaning, content, and possibility of events. \citet{pedinotti2021did} found that models may lack crucial aspects of human event knowledge and necessitate the ability to deduce the necessary physical properties for an object's involvement in an event. \citet{ding2019event} learns event representation by understanding its sentiment and intent.

\noindent\textbf{Reasoning} \citet{du2022care} proposed \texttt{ECARE} containing more than 13,000 distinct conceptual explanations regarding the profound comprehension of causal facts. \citet{zhou2020temporal} proposed a distant supervision process that enhances the reasoning of language models regarding the initiation times of both explicit and implicit events. \citet{qin2019counterfactual, qin2020back} presented a challenging scenario for current language understanding and generation systems by incorporating the need for counterfactual reasoning. \citet{han2021ester} proposed an MRC dataset for comprehensive event semantic reasoning. 

\noindent\textbf{Prediction} \citet{granroth2016happens} presented a narrative cloze evaluation task that is specifically designed for scenarios where abundant event information is available. \citet{lv2019sam} introduced a self-attention mechanism that effectively directs attention to various event segments. They also employed event-level attention to capture the relationships between subsequent events and individual events. \citet{bai2022rich} introduced the Rich Event Prediction framework for event prediction. The core concept of this framework revolves around enhancing existing descriptions by incorporating verb senses, additional semantic roles, and participant types.
\section{Conclustion}
In this paper, we present a comprehensive framework for event semantic processing that encompasses understanding, reasoning, and prediction, including their nuanced aspects. To thoroughly assess the event semantic processing capabilities of models, we introduce a novel benchmark named \ee, which comprises eight datasets covering various facets of event semantics. Through extensive experimentation on \ee, we uncover findings based on the obtained results including:

\begin{itemize}[topsep=0pt]
\setlength{\itemsep}{1pt}
\setlength{\parskip}{0pt}
\setlength{\parsep}{-1pt}
\setlength{\leftmargin}{-1pt}
\item[1] Although LLMs possess an understanding of individual events, their capacity to perceive the semantic similarity among events is constrained.
    
\item[2] LLMs exhibit robust reasoning abilities in causal and intentional relations, yet their performance in other relation types is comparatively weaker.
    
\item[3] With increased contextual information, LLMs exhibit enhanced predictive capabilities for future events.

\item[4] The effective utilization of event knowledge, context, and patterns holds importance in improving the CoT of event semantic processing.

\item[5] Structural event representation demonstrates performance on par with natural language in the domain of event processing.

\end{itemize}

\bibliography{main}
\bibliographystyle{acl_natbib}

\appendix

\end{document}